\documentclass{article}
\usepackage{arxiv}

\usepackage[utf8]{inputenc} 
\usepackage[T1]{fontenc}    
\usepackage{hyperref}       
\usepackage{url}            
\usepackage{booktabs}       
\usepackage{amsfonts}       
\usepackage{nicefrac}       
\usepackage{microtype}      
\usepackage{lipsum}		    
\usepackage{graphicx}
\usepackage{doi}
\usepackage{multirow}
\usepackage{multicol}
\usepackage{xcolor}
\usepackage{listings}
\usepackage{subcaption} 
\usepackage[normalem]{ulem}
\usepackage{amsmath,amsfonts,bm}
\usepackage{amssymb,amsthm}
\usepackage{enumitem}
\usepackage{algorithm}
\usepackage{algorithmic}
\usepackage{footnote}
\makesavenoteenv{tabular}
\usepackage[
    backend=biber,
    citestyle=authoryear,
    bibstyle=authortitle,
    mincitenames=1,
    maxcitenames=1,
    maxbibnames=3,
]{biblatex}
\newcommand{\citep}[1]{\parencite{#1}}
\newcommand{\ours}{{Moonlight}}

\usepackage{listings}
\newtheoremstyle{italicstyle} 
  {3pt} 
  {3pt} 
  {\itshape} 
  {} 
  {\itshape} 
  {.} 
  {.5em} 
  {} 
\theoremstyle{italicstyle}
\newtheorem{lemma}{Lemma}

\definecolor{darkblue}{rgb}{0.0, 0.0, 0.5}
\definecolor{darkgreen}{rgb}{0.0, 0.5, 0.0}
\definecolor{darkred}{rgb}{0.5, 0.0, 0.0}
\definecolor{darkpurple}{rgb}{0.5, 0.0, 0.5}

\setlist[itemize,1]{leftmargin=\dimexpr 18pt}
\setlist[enumerate,1]{leftmargin=\dimexpr 18pt}

\title{
\raisebox{-0.1\height}{\includegraphics[width=0.032\textwidth]{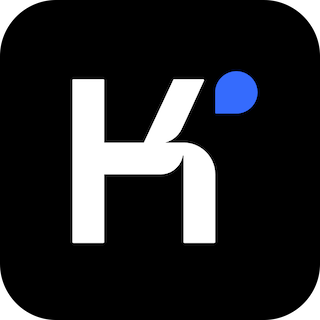}} %
Muon is Scalable for LLM Training
}

\author{Kimi Team}

\date{}

\renewcommand{\headeright}{Technical Report}
\renewcommand{\undertitle}{Technical Report}
\renewcommand{\shorttitle}{\raisebox{-0.12\height}{\includegraphics[width=0.02\textwidth]{figures/logo.png}}
Muon is Scalable for LLM Training}

\author{%
    \textbf{Jingyuan Liu}$^{1}$ \quad \textbf{Jianlin Su}$^{1}$ \quad \textbf{Xingcheng Yao}$^{2}$ \quad \textbf{Zhejun Jiang}$^{1}$ \quad \textbf{Guokun Lai}$^{1}$ \quad \textbf{Yulun Du}$^{1}$ \\
    \textbf{Yidao Qin}$^{1}$ \quad \textbf{Weixin Xu}$^{1}$ \quad \textbf{Enzhe Lu}$^{1}$ \quad \textbf{Junjie Yan}$^{1}$ \quad \textbf{Yanru Chen}$^{1}$ \quad \textbf{Huabin Zheng}$^{1}$ \\ \quad \textbf{Yibo Liu}$^{1}$ 
    \quad \textbf{Shaowei Liu}$^{1}$ \quad \textbf{Bohong Yin}$^{1}$ \quad \textbf{Weiran He}$^{1}$ \quad \textbf{Han Zhu}$^{1}$ \quad \textbf{Yuzhi Wang}$^{1}$ \quad \\ \textbf{Jianzhou Wang}$^{1}$ 
    \textbf{Mengnan Dong}$^{1}$ \quad \textbf{Zheng Zhang}$^{1}$ \quad \textbf{Yongsheng Kang}$^{1}$ \quad \textbf{Hao Zhang}$^{1}$ \quad \\ \textbf{Xinran Xu}$^{1}$ 
    \quad \textbf{Yutao Zhang}$^{1}$ \quad \textbf{Yuxin Wu}$^{1}$  \quad \textbf{Xinyu Zhou}$^{1}$ \thanks{Corresponding author: \texttt{zhouxinyu@moonshot.cn}} \quad \textbf{Zhilin Yang}$^{1} $
    \\[2ex]
    $^1$ Moonshot AI \quad $^2$ UCLA \quad
}

\begin{document}
\maketitle

\vspace{-10pt}
\begin{abstract}

Recently, the Muon optimizer~\citep{jordan2024muon} based on matrix orthogonalization has demonstrated strong results in training small-scale language models, but the scalability to larger models has not been proven. We identify two crucial techniques for scaling up Muon: (1) adding weight decay and (2) carefully adjusting the per-parameter update scale. These techniques allow Muon to work out-of-the-box on large-scale training without the need of hyper-parameter tuning. Scaling law experiments indicate that Muon achieves $\sim\!2\times$ computational efficiency compared to AdamW with compute optimal training.
Based on these improvements, we introduce \ours, a 3B/16B-parameter Mixture-of-Expert (MoE) model trained with 5.7T tokens using Muon. Our model improves the current Pareto frontier, achieving better performance with much fewer training FLOPs compared to prior models.
We open-source our distributed Muon implementation that is memory optimal and communication efficient. We also release the pretrained, instruction-tuned, and intermediate checkpoints to support future research.

\end{abstract}

\begin{figure}[h]
    \centering
    \subfloat[]{
        \includegraphics[width=0.446\textwidth]{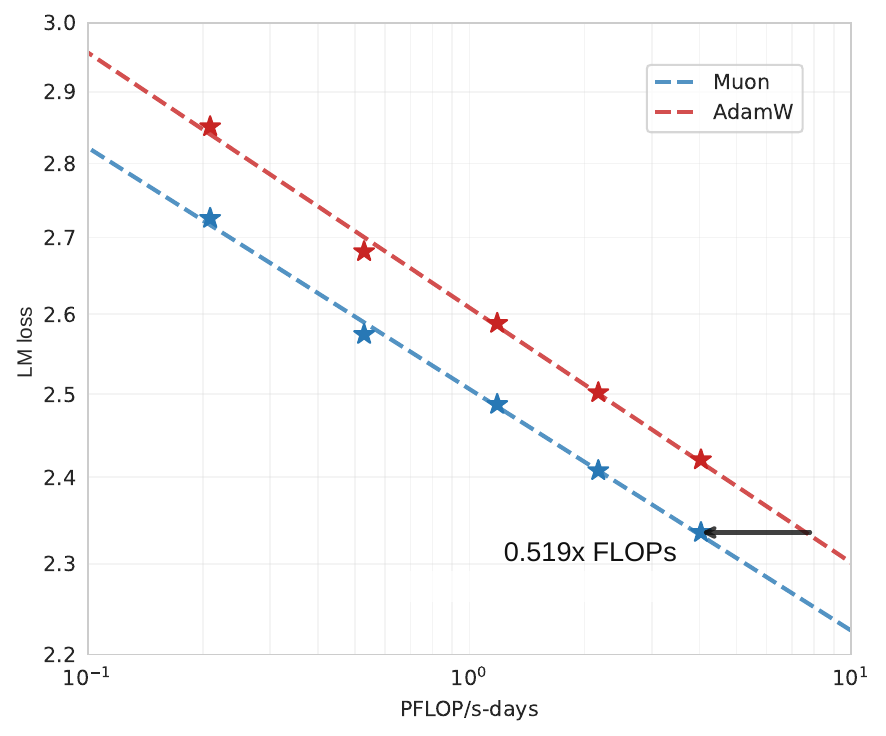}
        \label{fig:scaling_lm_loss}
        \vspace{-0.125cm} 
    }
    \subfloat[]{
        \includegraphics[width=0.53\textwidth]{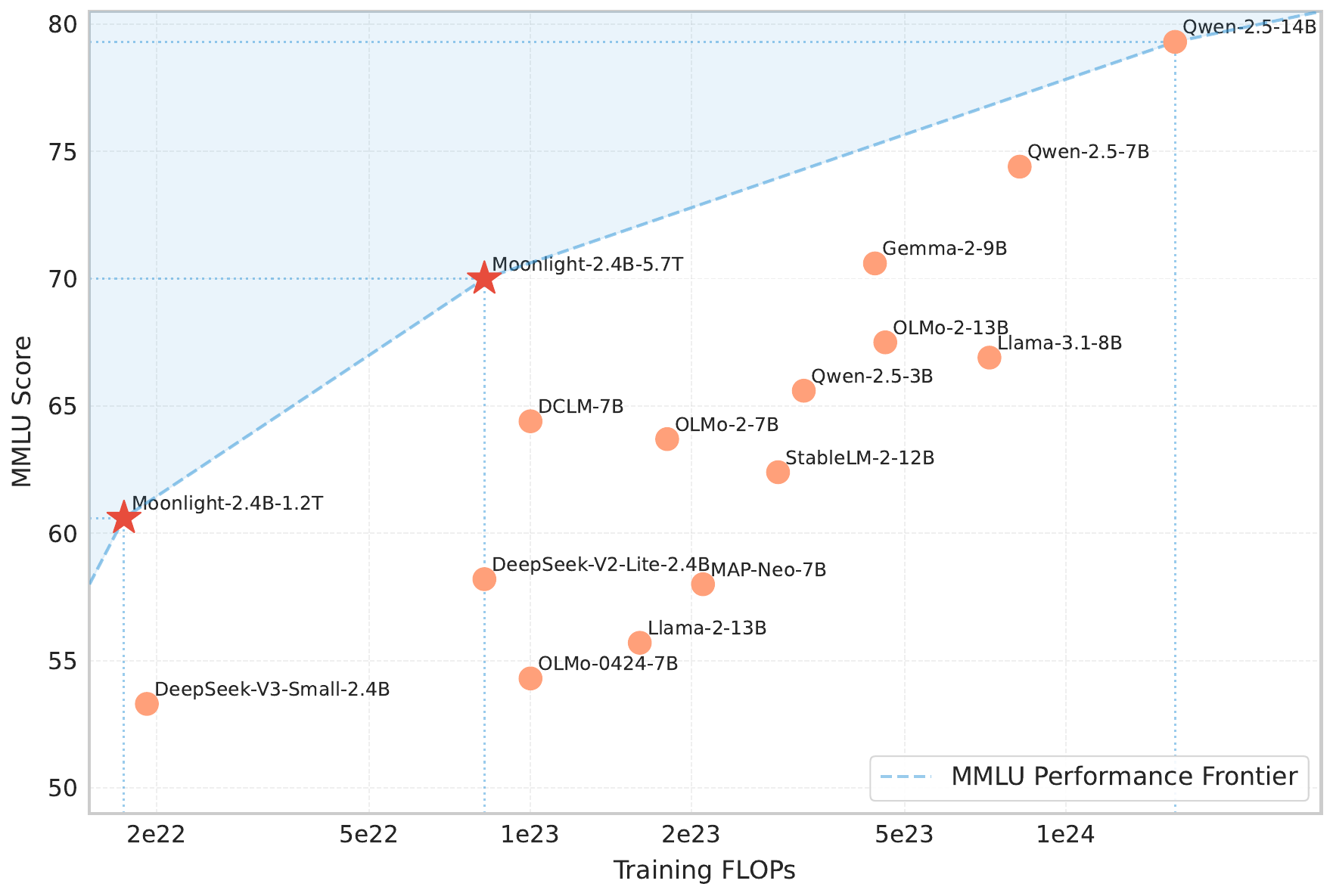}
        \label{fig:mmlu}
    }
    \caption{Scaling up with Muon. \textbf{(a)} Scaling law experiments comparing Muon and Adam. Muon is $\sim2\times$ more computational efficient than Adam with compute optimal training. \textbf{(b)} The MMLU performance of our Moonlight model optimized with Muon and other comparable models. Moonlight advances the Pareto frontier of performance vs training FLOPs.} 
    \label{fig:teaser} 
\end{figure}

\section{Introduction}

The rapid advancement of large language models (LLMs)~\citep{openai2024gpt4technicalreport,deepseekai2024deepseekv3technicalreport,grattafiori2024llama3herdmodels,geminiteam2024geminifamilyhighlycapable} has significantly pushed forward the progress in artificial general intelligence. However, training capable LLMs remains a computationally intensive and resource-demanding process due to scaling laws~\citep{kaplan2020scalinglawsneurallanguage,hoffmann2022trainingcomputeoptimallargelanguage}. Optimizers play a crucial role in efficiently and effectively training of LLMs, with Adam~\citep{adam2015kingma} and its variant AdamW~\citep{loshchilov2018decoupled} being the standard choice for most large-scale training.

Recent developments in optimization algorithms have shown potential to improve training efficiency beyond AdamW~\citep{liu2024sophia,jordan2024muon,yuan2024mars,vyas2025soap,Li_2018,li2018preconditionermatrixliegroup,pooladzandi2024curvatureinformedsgdgeneralpurpose,li2022blackboxliegroup,li2024stochastichessianfittingslie,pethick2025trainingdeeplearningmodels}. Among these, \cite{jordan2024muon} proposed Muon, which updates matrix parameters with orthogonalized gradient momentum using Newton-Schulz iteration. Initial experiments with Muon have demonstrated promising results in small-scale language model training. However, as discussed in this blog \citep{jordan2024muon}, several critical challenges remain unaddressed: (1) how to effectively scale optimizers based on matrix orthogonalization to larger models with billions of parameters trained with trillions of tokens, (2) how to compute approximate orthogonalization in a distributed setting, and (3) whether such optimizers can generalize across different training stages including pre-training and supervised finetuning (SFT).

In this technical report, we present a comprehensive study addressing these challenges. Our work builds upon Muon while systematically identifying and resolving its limitations in large-scale training scenarios. Our technical contributions include:

\begin{itemize}
    \item \textbf{Analysis for Effective Scaling of Muon}: Through extensive analysis, we identify that weight decay plays a crucial role in Muon's scalability. Besides, we propose scale adjustments to Muon's parameter-wise update rule. Such adjustments allow Muon to work out-of-the-box without hyper-parameter tuning, and also significantly improve training stability.
    
    \item \textbf{Efficient Distributed Implementation}: We develop a distributed version of Muon with ZeRO-1~\citep{Rajbhandari_2020} style optimization, achieving optimal memory efficiency and reduced communication overhead while preserving the mathematical properties of the algorithm.
    
    \item \textbf{Scaling Law Validation}: We performed scaling law research that compares Muon with strong AdamW baselines, and showed the superior performance of Muon (\ref{fig:scaling_lm_loss}). Based on the scaling law results, Muon achieves comparable performance to AdamW trained counterparts while requiring only approximately 52\% of the training FLOPs.

\end{itemize}

Our comprehensive experiments demonstrate that Muon can effectively replace AdamW as the de facto optimizer for large-scale LLM training, offering significant improvements in both training efficiency and model performance. As a result of this work, we release \ours, a 16B-parameter MoE model trained using Muon, along with our implementation and intermediate training checkpoints to facilitate further research in scalable optimization techniques for LLMs. 
\section{Methods}

\subsection{Background}

\paragraph{The Muon Optimizer}
\label{sec:analysis:background}
Muon~\citep{jordan2024muon} has recently been proposed to optimize neural network weights representable as matrices. At iteration $t$, given current weight $\mathbf{W}_{t-1}$, momentum $\mu$, learning rate $\eta_t$ and objective $\mathcal{L}_t$, the update rule of the Muon optimizer can be stated as follows:
\begin{align}
    \mathbf{M}_t &= \mu \mathbf{M}_{t-1} + \nabla\mathcal{L}_t(\mathbf{W}_{t-1}) \notag \\
    \mathbf{O}_t &= \text{Newton-Schulz}(\mathbf{M}_t)\text{\footnotemark[1]} \label{eq:Ot}\\
    \mathbf{W}_t &= \mathbf{W}_{t-1} - \eta_t \mathbf{O}_t \notag
\end{align}
 Here, $\mathbf{M}_t$ is the momentum of gradient at iteration $t$, set as a zero matrix when $t = 0$. In Equation~\ref{eq:Ot}, a Newton-Schulz iteration process~\citep{bernstein2024oldoptimizernewnorm} is adopted to approximately solve $(\mathbf{M}_t \mathbf{M}^{\mathrm{T}}_t)^{-1/2}\mathbf{M}_t$\footnotetext[1] {In practice, we follow~\citep{jordan2024muon} to use a Nesterov-style momentum by putting $\mu \mathbf{M}_t + \nabla\mathcal{L}_t(\mathbf{W}_{t-1})$ to the Newton-Schulz iteration instead of $\mathbf{M}_t$.}. Let $\mathbf{U}\mathbf{\Sigma} \mathbf{V}^\mathrm{T} = \mathbf{M}_t$ be the singular value decomposition (SVD) of $\mathbf{M}_t$, we will have $(\mathbf{M}_t \mathbf{M}^{\mathrm{T}}_t)^{-1/2}\mathbf{M}_t = \mathbf{U}\mathbf{V^T}$, which orthogonalizes $\mathbf{M}_t$. Intuitively, orthogonalization can ensure that the update matrices are isomorphic, preventing the weight from learning along a few dominant directions~\citep{jordan2024muon}.

\paragraph{Newton-Schulz Iterations for Matrix Orthogonalization}
Equation~\ref{eq:Ot} is calculated in an iterative process. At the beginning, we set $\mathbf{X}_0 = \mathbf{M}_t / \|\mathbf{M}_t\|_\mathrm{F}$. Then, at each iteration $k$, we update $\mathbf{X}_k$ from $\mathbf{X}_{k-1}$ as follows:
\begin{align}
    \mathbf{X}_k &= a \mathbf{X}_{k-1} + b (\mathbf{X}_{k-1} \mathbf{X}_{k-1}^\mathrm{T}) \mathbf{X}_{k-1} + c (\mathbf{X}_{k-1} \mathbf{X}_{k-1}^\mathrm{T})^2 \mathbf{X}_{k-1} \label{eq:iteration}
\end{align}
where $\mathbf{X}_N$ is the result of such process after $N$ iteration steps.
Here $a$, $b$, $c$ are coefficients. In order to ensure the correct convergence of Equation~\ref{eq:iteration}, we need to tune the coefficients so that the polynomial $f(x) = a x + b x^3 + c x^5$ has a fixed point near 1. In the original design of \cite{jordan2024muon}, the coefficients are set to $a = 3.4445$, $b = -4.7750$, $c = 2.0315$ in order to make the iterative process converge faster for small initial singular values. In this work, we follow the same setting of coefficients.

\paragraph{Steepest Descent Under Norm Constraints}
\cite{bernstein2024oldoptimizernewnorm} proposed to view the optimization process in deep learning as steepest descent under norm constraints. From this perspective, we can view the difference between Muon and Adam~\citep{adam2015kingma, loshchilov2018decoupled} as the difference in norm constraints. Whereas Adam is a steepest descent under the a norm constraint dynamically adjusted from a Max-of-Max norm, Muon offers a norm constraint that lies in a static range of Schatten-$p$ norm for some large $p$~\citep{muoncase2024cesista}. When equation~\ref{eq:Ot} is accurately computed, the norm constraint offered by Muon will be the spectral norm. Weights of neural networks are used as operators on the input space or the hidden space, which are usually (locally) Euclidean~\citep{cesista2024firstordernormedopt}, so the norm constraint on weights should be an induced operator norm (or spectral norm for weight matrices). In this sense, the norm constraint offered by Muon is more reasonable than that offered by Adam.

\subsection{Scaling Up Muon}
\label{sec:analysis:rms}

\paragraph{Weight Decay}

While Muon performs significantly better than AdamW on a small scale as shown by \cite{jordan2024muon}, we found the performance gains diminish when we scale up to train a larger model with more tokens. We observed that both the weight and the layer output's RMS keep growing to a large scale, exceeding the high-precision range of bf16, which might hurt the model's performance. To resolve this issue, we introduced the standard AdamW (\cite{loshchilov2018decoupled}) weight decay mechanism into Muon\footnote{The original implementation of Muon omits weight decay. A recent concurrent work in Muon incorporates weight decay and demonstrates improved performance. See \href{https://github.com/KellerJordan/Muon/commit/e0ffefd4f7ea88f2db724caa2c7cfe859155995d}{this commit} and \href{https://x.com/kellerjordan0/status/1888320690543284449}{this discussion}.}.

\begin{align}
\label{equation:weightdecay}
    \mathbf{W}_t = \mathbf{W}_{t-1} - \eta_t (\mathbf{O}_t + \lambda \mathbf{W}_{t-1})
\end{align}

\begin{figure}[t]
    \centering
    \includegraphics[width=0.8\textwidth]{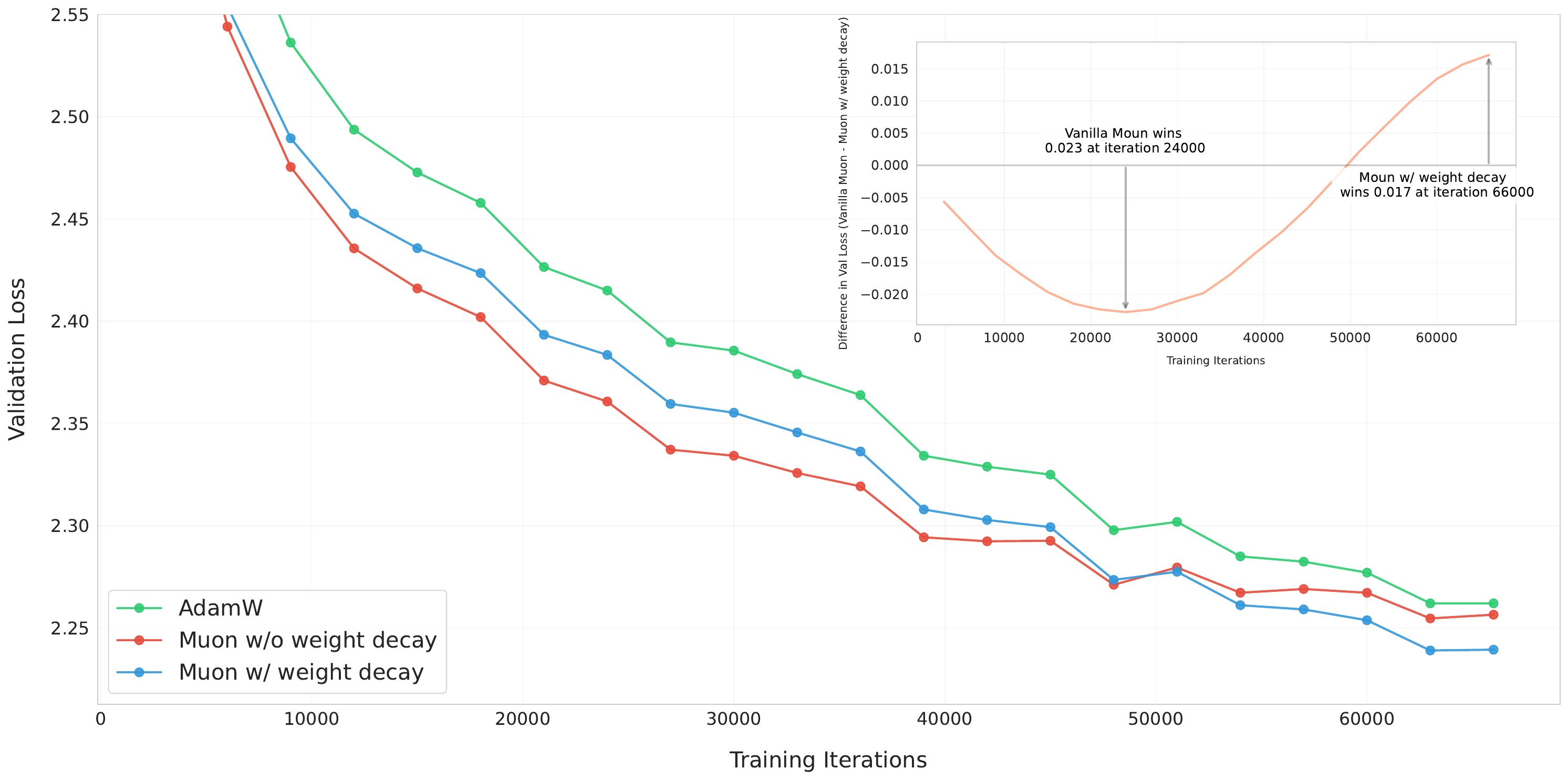}
    \caption{\small Validation loss curves for AdamW (\textcolor[HTML]{2ecc71}{green}), Muon without weight decay (\textcolor[HTML]{e74c3c}{red}), and Muon with weight decay (\textcolor[HTML]{3498db}{blue}).} 
    \label{fig_weight_decay} 
\end{figure}

We experimented on Muon both with and without weight decay to understand its impact on the training dynamics of LLMs. Based on our scaling law research in Sec \ref{sec:exp:moonscalinglaw}, we trained an 800M parameters model with 100B tokens ($\sim5\times$ optimal training tokens). Figure \ref{fig_weight_decay} shows validation loss curves of the model trained with AdamW, vanilla Muon (without weight decay), and Muon with weight decay. While vanilla Muon initially converges faster, we observed that some model weights grew too large over time, potentially limiting the model's long-term performances. Adding weight decay addressed this issue - the results demonstrate that Muon with weight decay outperforms both vanilla Muon and AdamW, achieving lower validation loss in the over-train regime. Therefore, we adjusted our update rule to equation \ref{equation:weightdecay}, where $\lambda$ is the weight decay ratio.

\paragraph{Consistent update RMS}
An important property of Adam and AdamW (\cite{adam2015kingma}, \cite{loshchilov2018decoupled}) is that they maintain a theoretical update RMS around 1\footnote{Due to Adam's $\beta_1 < \beta_2$ and $\epsilon > 0$, the actual update RMS is usually less than 1.}. However, we show that Muon's update RMS varies depending on the shape of the parameters, according to the following lemma:

\begin{lemma}
\label{lemma:updaterms}
For a full-rank matrix parameter of shape $[A, B]$, its theoretical Muon update RMS is $\sqrt{1/\max(A,B)}$ .
\end{lemma}

The proof can be found in the Appendix \ref{sec:appendix:updaterms}. We monitored Muon's update RMS during training and found it typically close to the theoretical value given above. We note that such inconsistency can be problematic when scaling up the model size:

\begin{itemize}
    \item When $\max(A,B)$ is too large, e.g. the dense MLP matrix, the updates become too small, thus limiting the model's representational capacity and leading to suboptimal performances; 
    
    \item When $\max(A,B)$ is too small, e.g. treating each KV head in GQA (\cite{shazeer2019fasttransformerdecodingwritehead}) or MLA (\cite{deepseekai2024deepseekv3technicalreport}) as a separate parameter, the updates become too large, thus causing training instabilities and leading to suboptimal performances as well.
\end{itemize}

In order to maintain consistent update RMS among matrices of different shapes, we 
propose to scale the Muon update for each matrix by its $\sqrt{\max(A, B)}$ to cancel the effect of Lemma~\ref{lemma:updaterms} \footnote{\cite{jordan2024muon}'s original implementation scales the updates by $\sqrt{\max(1, A/B)}$, which is equivalent to our proposal (up to a global scale) if all matrices have the same second dimension; \cite{pethick2025trainingdeeplearningmodels} and \cite{JiachengX} discussed a similar issue on update scaling factors concurrently to our work. } . 
Experiments in Sec~\ref{sec:exp:rms} show that this strategy is beneficial for optimization.

\paragraph{Matching update RMS of AdamW}

Muon is designed to update matrix-based parameters. In practice, AdamW is used in couple with Muon to handle non-matrix based parameters, like RMSNorm, LM head, and embedding parameters. 
We would like the optimizer hyper-parameters (learning rate $\eta$, weight decay $\lambda$) to be shared among
matrix and non-matrix parameters. 

We propose to match Muon's update RMS to be similar to that of AdamW. From empirical observations, AdamW's update RMS is usually around 0.2 to 0.4. Therefore, we scale Muon's update RMS to this range by the following adjustment:

\begin{align}
\mathbf{W}_t = \mathbf{W}_{t-1} - \eta_t (0.2\cdot\mathbf{O}_t\cdot\sqrt{\max(A,B)} + \lambda \mathbf{W}_{t-1})
\end{align}

 We validated this choice with empirical results (see Appendix \ref{sec:appendix:updaterms} for details). 
Moreover, we highlighted that with this adjustment, Muon can directly \textbf{reuse} the learning rate and weight decay tuned for AdamW. 

\paragraph{Other Hyper-parameters} Muon contains two other tunnable hyper-parameters: Newton-Schulz iteration steps and momentum $\mu$. We empirically observe that when setting $N$ to $10$, the iterative process will yield a more accurate orthogonalization result than $N=5$, but it won't lead to better performances. Hence we set $N = 5$ in this work for the sake of efficiency. We do not see a consistent performance gain in tuning momentum, so we chose 0.95, same as \cite{jordan2024muon}.

\subsection{Distributed Muon}
\label{sec:analysis:distrib}

\paragraph{ZeRO-1 and Megatron-LM}
\cite{Rajbhandari_2020} introduced the ZeRO-1 technique that partitions the expensive optimizer states (e.g. master weights, momentum) all over the cluster. Megatron-LM \citep{shoeybi2020megatronlmtrainingmultibillionparameter} integrated ZeRO-1 into its native parallel designs. Based on Megatron-LM's sophisticated parallel strategies, e.g. Tensor-Parallel (TP), Pipeline Parallel (PP), Expert Parallel (EP) and Data Parallel (DP), the communication workload of ZeRO-1 can be reduced from gathering all over the distributed world to only gathering over the data parallel group.

\paragraph{Method}
ZeRO-1 is efficient for AdamW because it calculates updates in an element-wise fashion. However, Muon requires the full gradient matrix to calculate the updates. Therefore, vanilla ZeRO-1 is not directly applicable to Muon. We propose a new distributed solution based on ZeRO-1 for Muon, referred to as Distributed Muon. Distributed Muon follows ZeRO-1 to partition the optimizer states on DP, and introduces two additional operations compared to a vanilla Zero-1 AdamW optimizer:

\begin{enumerate}
    \item \texttt{DP Gather.} For a local DP partitioned master weight ($1/DP$ the size of the model weight), this operation is to gather the corresponding partitioned gradients into a full gradient matrix. 
    
    \item \texttt{Calculate Full Update.} After the above gathering, perform Newton-Schulz iteration steps on the full gradient matrix as described in Sec \ref{sec:analysis:background}. Note that we will then discard part of the full update matrix, as we only need the partition corresponding to the local parameters to perform update.
\end{enumerate}

The implementation of Distributed Muon is described in Algorithm \ref{alg:distribmuon}. The additional operations introduced by Distributed Muon are colored in blue.

\begin{algorithm}[t]
\caption{Distributed Muon}
\label{alg:distribmuon}
\begin{algorithmic}[1]
\REQUIRE{Full Gradients $\mathbf{G}$, DP partitioned Momentum $\mathbf{m}$, DP partitioned parameters $\mathbf{p}$, momentum $\mu$.}
\STATE // Reduce-scatter $G$ on DP for correct gradients
\STATE $\mathbf{g} = \text{reduce\_scatter($\mathbf{G}$, dp\_group)}$ 
\STATE // Apply momentum to $\mathbf{g}$   using local partitioned momentum $\mathbf{m}$
\STATE $\mathbf{g}' = \text{update\_with\_momentum}(\mathbf{g}, \mathbf{m}, \mu)$
\STATE \textcolor{blue}{// DP Gather: gathering $\mathbf{g'}$ across DP into a full matrix $\mathbf{G}$}
\STATE \textcolor{blue}{$\mathbf{G} = \text{gather($\mathbf{g'}$, dp\_group)}$}
\STATE \textcolor{blue}{// Calculate Muon update}
\STATE \textcolor{blue}{$\mathbf{U} = \text{Newton-Schulz}(\mathbf{G})$ }
\STATE \textcolor{blue}{// Discard the rest of $\mathbf{U}$ and only keep the local partition  ${\mathbf{u}}$, then apply the update rule}
\STATE $\mathbf{p}' = \text{apply\_update}(\mathbf{p}, \mathbf{u})$
\STATE // All-gather updated $\mathbf{p'}$ into $\mathbf{P}$ 
\STATE $\mathbf{P} = \text{all\_gather($\mathbf{p'}$, dp\_group)}$
\STATE // Return the update RMS for logging
\RETURN $\sqrt{\mathbf{u}^2.\texttt{mean}()}$ 
\end{algorithmic}
\end{algorithm}

\paragraph{Analysis}
We compared Distributed Muon to a classic ZeRO-1 based distributed AdamW (referred as Distributed AdamW for simplicity) in several aspects:

\begin{itemize}
\item \texttt{Memory Usage.} Muon uses only one momentum buffer, while AdamW uses two momentum buffers. Therefore, the additional memory used by the Muon optimizer is half of Distributed AdamW.

\item \texttt{Communication Overhead.} For each device, the additional DP gathering is only required by the local DP partitioned parameters $\mathbf{p}$. Therefore, the communication cost is less than the reduce-scatter of $\mathbf{G}$ or the all-gather of $\mathbf{P}$. Besides, Muon only requires the Newton-Schulz iteration steps in bf16, thus further reducing the communication overhead to 50\% comparing to fp32. Overall, the communication workload of Distributed Muon is $(1, 1.25]$ of that of Distributed AdamW. The upper-bound is calculated as that the communication of Distributed Muon is 4 (fp32 $\mathbf{G}$ reduce-scatter) + 2 (bf16 Muon gather) + 4 (fp32 $\mathbf{P}$ all-gather), while Distributed AdamW is 4 + 4. In practice, as we usually train with multiple DP, the empirical additional cost usually is closer to the lower-bound 1.\footnote{If TP is enabled, Distributed Muon needs an extra bf16 TP gather on TP group.}.

\item \texttt{Latency.} Distributed Muon has larger end-to-end latencies than Distributed AdamW because it introduces additional communication and requires running Newton-Schulz iteration steps. However, this is not a significant issue because (a) only about 5 Newton-Schultz iteration steps are needed for a good result (discussed in Sec \ref{sec:analysis:rms}), and (b) the end-to-end latency caused by the optimizer is negligible compared to the model's forward-backward pass time (e.g. usually 1\% to 3\%). Moreover, several engineering techniques, such as overlapping gather and computation, and overlapping optimizer reduce-scatter with parameter gather, can further reduce latency.

\end{itemize}

When training large-scale models in our distributed cluster, Distributed Muon has no noticeable latency overhead compared to its AdamW counterparts. We will soon release a pull request that implements Distributed Muon for the open-source Megatron-LM \citep{shoeybi2020megatronlmtrainingmultibillionparameter} project.

\section{Experiments}

\subsection{Consistent Update RMS}
\label{sec:exp:rms}

As discussed in Sec \ref{sec:analysis:rms}, we aim to match the update RMS across all matrix parameters and also match it with that of AdamW. We experimented with two methods to control the Muon update RMS among parameters and compared them to a baseline that only maintains a consistent RMS with AdamW:

\begin{enumerate}
    \item \texttt{Baseline.} We multiplied the update matrix by $0.2\cdot \sqrt{H}$ ($H$ is the model hidden size) to maintain a consistent update RMS with AdamW. Note that $\max(A,B)$ equals to $H$ for most matrices.
    \begin{align}
    \mathbf{W}_t = \mathbf{W}_{t-1} - \eta_t (0.2\cdot\mathbf{O}_t\cdot\sqrt{H} + \lambda \mathbf{W}_{t-1})
    \end{align}
    \item \texttt{Update Norm.} We can directly normalize the updates calculated via Newton-Schulz iterations so its RMS strictly becomes 0.2;
    \begin{align}
    \mathbf{W}_t = \mathbf{W}_{t-1} - \eta_t (0.2\cdot\mathbf{O}_t/\mathop{\text{RMS}}(\mathbf{O}_t) + \lambda \mathbf{W}_{t-1})
    \end{align}
    \item \texttt{Adjusted LR.} For each update matrix, we can scale its learning rate by a factor of $0.2 \cdot \sqrt{\max(A, B)}$ based on its shape. 
    \begin{align}
    \mathbf{W}_t = \mathbf{W}_{t-1} - \eta_t (0.2\cdot\mathbf{O}_t\cdot\sqrt{\max(A,B)} + \lambda \mathbf{W}_{t-1})
    \end{align}
\end{enumerate}

\paragraph{Analysis}
We designed experiments to illustrate the impact of Muon update RMS at an early training stage, because we observed that unexpected behaviors happened very quickly when training models at larger scale. We experimented with small scale 800M models as described in \ref{sec:exp:moonscalinglaw}. The problem of inconsistent update RMS is more pronounced when the disparity between matrix dimensions increases. To highlight the problem for further study, we slightly modify the model architecture by replacing the Swiglu MLP with a standard 2-layer MLP, changing the shape of its matrix parameters from $[H, 2.6H]$ to $[H, 4H]$. We evaluated the model's loss and monitored a few of its parameters' RMS, specifically, attention query (shape $[H, H]$) and MLP (shape $[H, 4H]$). We evaluated the model after training for 4B tokens out of a 20B-token schedule. From Table~\ref{tab:muon-params-rms}, we observed several interesting findings:

\begin{table}[t]
\small
\centering
\caption{Controlling Muon's Update RMS Across Different Model Params}
\label{tab:muon-params-rms}
\begin{tabular}{c|c|c|c|c}
\toprule
Methods & Training loss & Validation loss & query weight RMS & MLP weight RMS \\
\midrule
Baseline & 2.734 & 2.812 & 3.586e-2 & 2.52e-2 \\
Update Norm & \textbf{2.72} & \textbf{2.789} & 4.918e-2 & 5.01e-2 \\
Adjusted LR & 2.721 & \textbf{2.789} & 3.496e-2 & 4.89e-2 \\
\bottomrule
\end{tabular}
\end{table}

\begin{enumerate}
    \item Both \texttt{Update Norm} and \texttt{Adjusted LR} achieved better performances than \texttt{Baseline};
    
    \item For the MLP weight matrix of shape $[H, 4H]$, both \texttt{Update Norm} and \texttt{Adjusted LR} obtain a weight RMS that is roughly doubled comparing to \texttt{Baseline}. This is reasonable as $\sqrt{\text{max}(H,4H)} / \sqrt{H} = 2$, so the update RMS of \texttt{Update Norm} and \texttt{Adjusted LR} is roughly two times of \texttt{Baseline};
    
    \item For the attention query weight matrix of shape $[H, H]$, \texttt{Update Norm} still norms the update, while \texttt{Adjusted LR} does not because $\sqrt{\text{max}(H,H)} / \sqrt{H} = 1$. As a result, \texttt{Adjusted LR} results in a similar weight RMS as \texttt{Baseline}, but \texttt{Update Norm} has a larger weight rms similar to its MLP.
\end{enumerate}

Based on these findings, we choose the \texttt{Adjusted LR} method for future experiments because it has lower cost.

\subsection{Scaling Law of Muon}
\label{sec:exp:moonscalinglaw}

For a fair comparison with AdamW, we performed scaling law experiments on a series of dense models in Llama \citep{grattafiori2024llama3herdmodels} architecture. Building a strong baseline is of crucial importance in optimizer research. Hence, we perform a grid search for hyper-parameters of AdamW, following the compute-optimal training setup \citep{kaplan2020scalinglawsneurallanguage} (the grid search experiments can be found in Appendix~\ref{sec:appendix:scaling}). Details of the model architecture and hyper-parameters can be found in Table~\ref{tab:model-specs}. For Muon, as discussed in Sec~\ref{sec:analysis:rms}, since we matched Muon's update RMS to AdamW, we directly reused the hyper-parameters that are optimal for the AdamW baseline.

\begin{table}[t]
\small
\centering
\caption{Scaling Law Models and Hyper-Parameters}
\label{tab:model-specs}
\begin{tabular}{c|c|c|c|c|c|c}
\toprule
\# Params. w/o Embedding & Head & Layer & Hidden & Tokens & LR & Batch Size* \\
\midrule
399M & 12 & 12 & 1536 & 8.92B  & 9.503e-4 & 96  \\
545M & 14 & 14 & 1792 & 14.04B & 9.143e-4 & 128 \\
822M & 16 & 16 & 2048 & 20.76B & 8.825e-4 & 160 \\
1.1B & 18 & 18 & 2304 & 28.54B & 8.561e-4 & 192 \\
1.5B & 20 & 20 & 2560 & 38.91B & 8.305e-4 & 256 \\
\bottomrule
\end{tabular}
\\ \footnotesize{\small *In terms of number of examples in 8K context length.} 
\end{table}

\begin{figure}[h]
    \centering
    \includegraphics[width=0.6\textwidth]{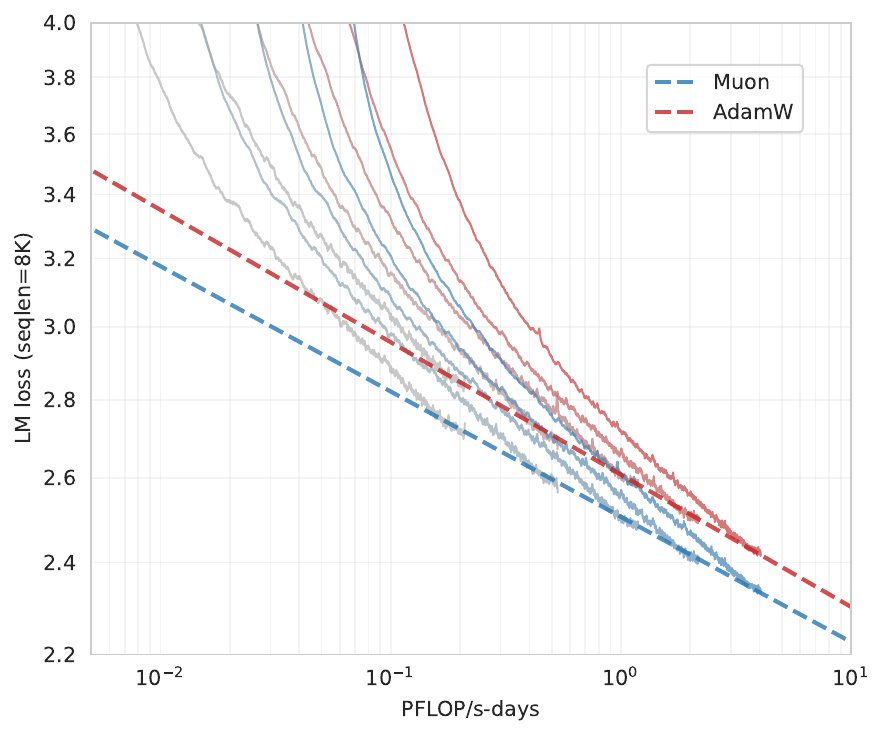}
    \caption{Fitted scaling law curves for Muon and AdamW optimizers.}
    \label{fig:scaling_lm_loss_fitting}
\end{figure}

The fitted scaling law curve can be found in figure \ref{fig:scaling_lm_loss_fitting}, and the fitted equations are detailed in table \ref{tab:fit}. As shown in Figure~\ref{fig:scaling_lm_loss}, Muon only requires about 52\% training FLOPs to match the performance of AdamW under compute-optimal setting.

\begin{table}
\centering
\caption{Fitted parameters of the scaling law curves}
\label{tab:fit}
\begin{tabular}{c|l|l}
\toprule
 & Muon & AdamW \\
\midrule
LM loss (seqlen=8K) & $2.506 \times C^{-0.052}$ & $2.608 \times C^{-0.054}$ \\
\bottomrule
\end{tabular}
\end{table}

\subsection{Pretraining with Muon}
\label{sec:exp:pretrain}

\paragraph{Model Architecture} To evaluate Muon against contemporary model architectures, we pretrained from scratch using the deepseek-v3-small architecture \citep{deepseekai2024deepseekv3technicalreport} as it demonstrates strong performance and the original results serve as a reference for comparison. Our pretrained model has 2.24B activated and 15.29B total parameters (3B activated and 16B total when including embedding). Minor modifications to the architecture are detailed in Appendix~\ref{sec:appendix:modelarch}.

\paragraph{Pretraining Data} Our pretraining data details can be found in \cite{k1p5}. The maximum context length during pretraining is 8K.

\paragraph{Pretraining} The model is trained in several stages. We use a 1e-3 auxfree bias update rate in stage 1 and 2, and 0.0 auxfree bias update rate in stage 3. The weight decay is set to 0.1 for all stages. More details and discussions of model training can be found in the Appendix \ref{sec:appendix:stability}.

\begin{enumerate}
    \item \texttt{0 to 33B tokens:} In this stage, the learning rate linearly increases to 4.2e-4 in 2k steps. The batch size is kept at 2048 examples;
    \item \texttt{33B to 5.2T tokens:} In this stage, the learning rate decays from 4.2e-4 to 4.2e-5 in a cosine style. We keep the batch size at 2048 until 200B tokens, and then doubled to 4096 for the remaining;
    \item \texttt{5.2T to 5.7T tokens:} In this stage (also referred as the cooldown stage), the learning rate increases to 1e-4 in in 100 steps, and then linearly decays to 0 in 500B tokens, and we keep a constant 4096 batch size. In this stage, we use the highest quality data, focusing on math, code, and reasoning.
\end{enumerate}

\paragraph{Evaluation Benchmarks} Our evaluation encompasses four primary categories of benchmarks, each designed to assess distinct capabilities of the model:

\begin{itemize}
    \item \textbf{English Language Understanding and Reasoning}: MMLU(5-shot)\citep{hendrycks2021measuringmassivemultitasklanguage}, MMLU-pro(5-shot) \citep{wang2024mmluprorobustchallengingmultitask}, BBH(3-shot) \citep{suzgun2022challengingbigbenchtaskschainofthought}, TriviaQA(5-shot) \citep{joshi2017triviaqalargescaledistantly}

    \item \textbf{Code Generation}: HumanEval(pass@1) \citep{chen2021codex}, MBPP(pass@1)\citep{austin2021programsynthesislargelanguage}
    
    \item  \textbf{Mathematical Reasoning}: GSM8K(4-shot) \citep{cobbe2021trainingverifierssolvemath} MATH \citep{hendrycks2021measuringmathematicalproblemsolving}, CMATH \citep{wei2023cmathlanguagemodelpass}

    \item \textbf{Chinese Language Understanding and Reasoning}: C-Eval(5-shot) \citep{huang2023cevalmultilevelmultidisciplinechinese}, CMMLU(5-shot)\citep{li2024cmmlumeasuringmassivemultitask}
\end{itemize}

\paragraph{Performance} We named our model trained with Muon ``Moonlight''. We compared Moonlight with different public models on a similar scale. We first evaluated Moonlight at 1.2T tokens and compared it with the following models that have the same architecture and trained with comparable number of tokens:

\begin{itemize}    
    \item \texttt{Deepseek-v3-Small } (\cite{deepseekai2024deepseekv3technicalreport}) is a  2.4B/16B-parameter MoE model trained with 1.33T tokens;
    \item \texttt{Moonlight-A} follows the same training settings as Moonlight, except that it uses the AdamW optimizer.
\end{itemize}

 For Moonlight and Moonlight-A, we used the intermediate 1.2T token checkpoint of the total 5.7T pretraining, where the learning rate is not decayed to minimal and the model has not gone through the cooldown stage yet.

\begin{table}[!ht]
    \small
    \centering
    \caption{Comparison of different models at around 1.2T tokens.}
    \setlength{\tabcolsep}{4pt}
    \begin{tabular}{@{}c l c c c c@{}}
    \toprule
    & \textbf{Benchmark (Metric)}  & \textbf{DSV3-Small} & \textbf{Moonlight-A@1.2T} & \textbf{Moonlight@1.2T} \\
    \midrule
    & Activated Params$^{\dagger}$ & 2.24B & 2.24B & 2.24B \\
    & Total Params$^{\dagger}$ & 15.29B & 15.29B & 15.29B \\
    & Training Tokens & 1.33T & 1.2T & 1.2T \\
    & Optimizer & AdamW & AdamW & Muon \\
    \midrule
    \multirow{4}{*}{English} 
    & MMLU & 53.3 & 60.2 & \textbf{60.4} \\
    & MMLU-pro & - & 26.8 & \textbf{28.1} \\
    & BBH & 41.4 & \textbf{45.3} & 43.2 \\
    & TriviaQA & -  & 57.4 & \textbf{58.1} \\
    \midrule
    \multirow{2}{*}{Code} & HumanEval & 26.8 & 29.3 & \textbf{37.2} \\
    & MBPP & 36.8 & 49.2 & \textbf{52.9} \\
    \midrule
    \multirow{3}{*}{Math} & GSM8K & 31.4 &  43.8 & \textbf{45.0} \\
    & MATH & 10.7 & 16.1 & \textbf{19.8} \\
    & CMath & - & 57.8 & \textbf{60.2} \\
    \midrule
    \multirow{2}{*}{Chinese} 
    & C-Eval & - &  57.2 & \textbf{59.9} \\
    & CMMLU & - & 58.2 & \textbf{58.8} \\
    \bottomrule
    \end{tabular}
    
    \footnotesize{\small $^{\dagger}$ The reported parameter counts exclude the embedding parameters.} 
    \label{tab:1.33Tresults}
\end{table}

As shown in Table \ref{tab:1.33Tresults}, Moonlight-A, our AdamW-trained baseline model, demonstrates strong performance compared to similar public models. Moonlight performs significantly better than Moonlight-A, proving the scaling effectiveness of Muon. We observed that Muon especially excels on Math and Code related tasks, and we encourage the research community to further investigate this phenomena. After Moonlight is fully trained to 5.7T tokens, we compared it with public models at similar scale and showed the results in Table \ref{tab:5.7Tresults_full}:

\begin{itemize}
    \item \texttt{LLAMA3-3B} from \cite{grattafiori2024llama3herdmodels} is a 3B-parameter dense model trained with 9T tokens. 
    \item \texttt{Qwen2.5-3B} from \cite{qwen2.5} is a 3B-parameter dense model trained with 18T tokens.
    \item \texttt{Deepseek-v2-Lite} from \cite{deepseekv2} is a 2.4B/16B-parameter MOE model trained with 5.7T tokens.
\end{itemize}

\begin{table}[!ht]
    \small
    \centering
    \caption{Comparison of different models on various benchmarks.}
    \setlength{\tabcolsep}{4pt}
    \begin{tabular}{@{}c l c c c c@{}}
    \toprule
    & \textbf{Benchmark (Metric)} & \textbf{Llama3.2-3B} & \textbf{Qwen2.5-3B} & \textbf{DSV2-Lite} & \textbf{Moonlight} \\
    \midrule
    & Activated Param$^{\dagger}$ & 2.81B & 2.77B & 2.24B & 2.24B \\
    & Total Params$^{\dagger}$ & 2.81B & 2.77B & 15.29B & 15.29B \\
    & Training Tokens  & 9T & 18T & 5.7T & 5.7T \\
    & Optimizer & AdamW  & Unknown & AdamW & Muon \\
    \midrule
    \multirow{4}{*}{English}
    & MMLU & 54.7 & 65.6 & 58.3 & \textbf{70.0} \\
    & MMLU-pro & 25.0 & 34.6 & 25.5 & \textbf{42.4} \\
    & BBH & 46.8 & 56.3 & 44.1 & \textbf{65.2} \\
    & TriviaQA$^{\ddagger}$ & 59.6 & 51.1 & 65.1 & \textbf{66.3} \\
    \midrule
    \multirow{2}{*}{Code} & HumanEval & 28.0 & 42.1 & 29.9 & \textbf{48.1} \\
    & MBPP & 48.7 & 57.1 & 43.2 & \textbf{63.8} \\
    \midrule
    \multirow{3}{*}{Math} & GSM8K & 34.0 & \textbf{79.1} & 41.1 & 77.4 \\
    & MATH & 8.5 & 42.6 & 17.1 & \textbf{45.3} \\
    & CMath & - & 80.0 & 58.4 & \textbf{81.1} \\
    \midrule
    \multirow{2}{*}{Chinese}
    & C-Eval & - & 75.0 & 60.3 & \textbf{77.2} \\
    & CMMLU & - & 75.0 & 64.3 & \textbf{78.2} \\
    \bottomrule
    \end{tabular}
    
    \footnotesize{$^{\dagger}$ The reported parameter counts exclude the embedding parameters.$^{\ddagger}$ We tested all listed models with the full set of TriviaQA.}
    \label{tab:5.7Tresults_full}
\end{table}

As shown in Table~\ref{tab:5.7Tresults_full}, Moonlight outperforms models with similar architectures trained with an equivalent number of tokens. Even when compared to dense models trained on substantially larger datasets, Moonlight maintains competitive performance. Detailed comparisons can be found in Appendix~\ref{sec:appendix:comparisons}. The performance of Moonlight is further compared with other well-known language models on MMLU and GSM8k, as illustrated in Figure~\ref{fig:mmlu} and Appendix~\ref{sec:appendix:comparisons} Figure~\ref{Fig:model_perf}.\footnote{Performance metrics and computational requirements (FLOPs) for baseline models are sourced from~\citep{olmo20242}}. Notably, Moonlight lies on the Pareto frontier of model performance versus training budget, outperforming many other models across various sizes.

\subsection{Dynamics of Singular Spectrum}
\begin{figure}[t]
    \centering
    \includegraphics[width=\textwidth]{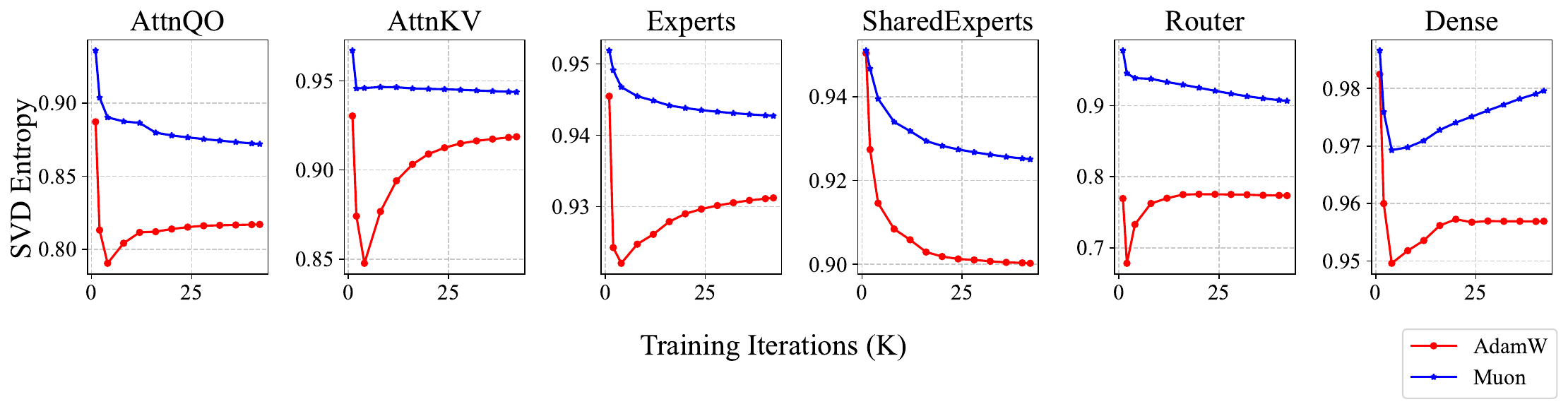}
    \caption{SVD entropy of weight matrices across different training iterations. We categorize the weight matrices into 6 different groups: 1) AttnQO denotes the weight matrices related to the query and output projection in the attention layer; 2) AttnKV denotes the weight matrices related to the key and value projection in the attention layer; 3) Experts denotes the weight matrices in expert models; 4) SharedExperts denotes the weight matrices in shared expert models; 5) Router denotes the weight matrices in the router; 6) Dense denotes the weight matrices in the first dense layer. The SVD entropy is calculated as the macro-average of the weight matrices in each group across all layers. For weights in expert models, we only calculate 3 out of 64 experts in different layers for efficiency.} 
    \label{fig_svd_entropy} 
\end{figure}

In order to validate the intuition that Muon can optimize the weight matrices in more diverse directions, we conducted a spectral analysis of the weight matrices trained with Muon and AdamW. For a weight matrix with singular values $\sigma = (\sigma_1, \sigma_2, \cdots, \sigma_n)$, we calculate the SVD entropy~\citep{svd_entropy, effectiverank} of this matrix as follows:
\begin{equation}
    H(\sigma) = -\frac{1}{\log n}\sum_{i=1}^n \frac{\sigma^2_i}{\sum_{j=1}^n \sigma^2_j} \log \frac{\sigma^2_i}{\sum_{j=1}^n \sigma^2_j} \notag
\end{equation}
As shown in Figure~\ref{fig_svd_entropy}, we visualized the average SVD entropy of the weight matrices across different training checkpoints during pretraining with 1.2T tokens. We can see that across all training checkpoints and all groups of weight matrices, the SVD entropy of Muon is higher than that of AdamW, which verifies the intuition that Muon can provide a more diverse spectrum of updates for the weight matrices. This discrepancy is more significant in the router weights for expert selection, which indicates that mixture-of-expert models can benefit more from Muon.

Moreover, we visualized the singular value distributions of each weight matrix at the checkpoint trained with 1.2T tokens as demonstrated in Appendix~\ref{sec:appendix:svd}. We find that, for over 90\% of the weight matrices, the SVD entropy when optimized by Muon is higher than that of AdamW, providing strong empirical evidence for Muon's superior capability in exploring diverse optimization directions.

\subsection{Supervised Finetuning (SFT) with Muon}

In this section, we present ablation studies on the Muon optimizer within the standard SFT stage of LLM training. Our findings demonstrate that the benefits introduced by Muon persist during the SFT stage. Specifically, a model that is both Muon-pretrained and Muon-finetuned outperforms others in the ablation studies. However, we also observe that when the SFT optimizer differs from the pretraining optimizer, SFT with Muon does not show a significant advantage over AdamW. This suggests that there is still considerable room for further exploration, which we leave for future work.

\subsubsection{Ablation Studies on the Interchangeability of Pretrain and SFT Optimizers}

To further investigate Muon’s potential, we finetuned Moonlight@1.2T and Moonlight-A@1.2T using both the Muon and AdamW optimizers. These models were finetuned for two epochs on the open-source tulu-3-sft-mixture dataset (\cite{lambert2024tulu3}), which contains 4k sequence length data. The learning rate followed a linear decay schedule, starting at $5 \times 10^{-5}$ and gradually reducing to $0$. The results, shown in Table \ref{tab:optim-interchangeability}, highlight the superior performance of Moonlight@1.2T compared to Moonlight-A@1.2T.

\begin{table}[ht]
\small
\centering
\caption{Examining the impact of optimizer interchangeability between pretraining and SFT phases.}
\label{tab:optim-interchangeability}
\begin{tabular}{l c|c|c|c|c}
\toprule
\textbf{Benchmark (Metric)} & \textbf{\# Shots} & \multicolumn{4}{|c}{\textbf{Moonlight-1.2T}} \\
\midrule
Pretraining Optimizer & - & Muon & AdamW & Muon & AdamW \\
SFT Optimzier & - & Muon & Muon & AdamW & AdamW \\
\midrule
MMLU (EM) & 0-shot (CoT) & \textbf{55.7} & 55.3 & 50.2 & 52.0 \\
HumanEval (Pass@1) & 0-shot & \textbf{57.3} & 53.7 & 52.4 & 53.1 \\
MBPP (Pass@1) & 0-shot & \textbf{55.6} & 55.5 & 55.2 & 55.2 \\
GSM8K (EM) & 5-shot & \textbf{68.0} & 62.1 & 64.9 & 64.6 \\
\bottomrule
\end{tabular}

\end{table}

\subsubsection{SFT with Muon on public pretrained models}

We further applied Muon to the supervised fine-tuning (SFT) of a public pretrained model, specifically the Qwen2.5-7B base model (\cite{qwen2.5}), using the open-source tulu-3-sft-mixture dataset (\cite{lambert2024tulu3}). The dataset was packed with an 8k sequence length, and we employed a cosine decay learning rate schedule, starting at $2 \times 10^{-5}$ and gradually decreasing to $2 \times 10^{-6}$. The results are presented in Table \ref{tab:public-model-SFT-results}. For comparison, we show that the Muon-finetuned model achieves performance on par with the Adam-finetuned model. These results indicate that for optimal performance, it is more effective to apply Muon during the pretraining phase rather than during supervised fine-tuning.

\begin{table}[ht]
\small
\centering
\caption{Comparison of Adam and Muon optimizers applied to the SFT of the Qwen2.5-7B pretrained model.}
\label{tab:public-model-SFT-results}
\begin{tabular}{l c|c|c}
\toprule
\textbf{Benchmark (Metric)} & \textbf{\# Shots} & \textbf{Adam-SFT} & \textbf{Muon-SFT} \\
\midrule
Pretrained Model & - & \multicolumn{2}{|c}{Qwen2.5-7B} \\
\midrule
MMLU (EM) & 0-shot (CoT) & \textbf{71.4} & 70.8 \\
HumanEval (Pass@1) & 0-shot & \textbf{79.3} & 77.4 \\
MBPP (Pass@1) & 0-shot & \textbf{71.9} & 71.6 \\
GSM8K (EM) & 5-shot & \textbf{89.8} & 85.8 \\
\bottomrule
\end{tabular}
\end{table}

\section{Discussions}

There are several possible directions for future research that could further explore and expand upon the current findings.

\paragraph{Incorporating All Parameters into the Muon Framework}
Currently, the Muon optimizer is utilized in conjunction with the Adam optimizer, where certain parameters remain under the purview of Adam optimization. This hybrid approach, while functional, presents an opportunity for improvement. The integration of the optimization of all parameters exclusively within the Muon framework is a topic of significant research interest.

\paragraph{Extending Muon to Schatten Norms}
The Muon optimizer can be interpreted as the steepest descent method under the spectral norm. Given the broad applicability and versatility of Schatten norms, extending Muon to encompass the general Schatten norm is a promising direction. This extension may unlock additional optimization capabilities and potentially yield superior results compared to the current spectral norm-based implementation.

\paragraph{Understanding and Solving the Pretraining-Finetuning Mismatch}
A notable phenomenon observed in practice is the suboptimal performance of models pretrained with AdamW when fine-tuned with Muon, and vice versa. This optimizer mismatch presents a significant barrier to effectively leveraging the extensive repository of AdamW-pretrained checkpoints, thereby necessitating a rigorous theoretical investigation. A precise understanding of the underlying mechanisms is essential for devising robust and effective solutions.

\section{Conclusions}
In this technical report, we presented a comprehensive study on the scalability of Muon in LLM training. Through systematic analysis and improvements, we successfully applied Muon to a 3B/16B-parameter MoE model trained on 5.7 trillion tokens. Our results demonstrate that Muon can effectively replace AdamW as the standard optimizer for large-scale LLM training, offering significant advantages in both training efficiency and model performance. By open-sourcing our implementation, the \ours~model, and intermediate training checkpoints, we aim to facilitate further research in scalable optimization techniques and accelerate the development of training methods for LLMs.
\newpage
\printbibliography
\newpage
\appendix
\section{Update RMS}
\label{sec:appendix:updaterms}

\paragraph{Proof of Lemma $\ref{lemma:updaterms}$}

\begin{proof}
Without loss of generality, consider the orthogonal matrices $U\in\mathbb{R}^{n\times n}$ and $V\in\mathbb{R}^{m\times m}$ where $n \geq m \geq r$. We will show that for $X=U_{[:,:r]}V_{[:r,:]}$ (the update of the Muon has the same format), the RMS value is $\sqrt{r/mn}$. From the definition of matrix multiplication:
$$X_{i,j}=\sum_{k=1}^r U_{i,k}V_{k,j}$$

The RMS can be expressed as:
$$\begin{aligned}
\text{RMS}(X)^2 &= \frac{1}{mn}\sum_{i=1}^n\sum_{j=1}^m \sum_{k=1}^r U_{i,k}^2V_{k,j}^2 \\
&= \frac{1}{mn}\sum_{k=1}^r\left(\sum_{i=1}^n U_{i,k}^2\right)\left(\sum_{j=1}^m V_{k,j}^2\right) \\
&= \frac{1}{mn}\sum_{k=1}^r 1 \\
&= \frac{r}{mn}
\end{aligned}$$

Therefore, $\text{RMS}(X)=\sqrt{r/mn}$. For the common case where the matrices are full-rank, $r=m$, yielding $\text{RMS}(X)=\sqrt{1/n}$.
\end{proof}

\paragraph{Consistent Update RMS Across Muon and AdamW}
As discussed in \ref{sec:analysis:rms}, we'd like to match the update RMS between Muon and AdamW optimizers. This is validated by experiments on small-scale models. We set Muon's Update RMS in the range of $[0.05, 0.1, 0.2, 0.4, 0.8]$ and AdamW as baseline. We reported the loss and representative weight matrix RMS at 2k steps (about 2B tokens) in the Table \ref{tab:muonrms}. From the results, we find that 0.2 RMS and 0.4 RMS performed similarly and much better than other settings. These findings are consistent with our empirical observation that AdamW's update RMS is in the range of $0.2\sim0.4$. We opted to control the update RMS of Muon to 0.2.

\begin{table}[ht]
\small
\centering
\caption{Muon Update RMS Experiments}
\label{tab:muonrms}
\begin{tabular}{c|c|c|c|c|c|c}
\toprule
Optimizer & AdamW & 0.05 RMS* & 0.1 RMS & 0.2 RMS & 0.4 RMS & 0.8 RMS\\
\midrule
LM training loss & 3.512 & 3.355 & 3.239 & \textbf{3.198} & 3.199 & 3.386  \\
LM validation loss & 3.679 & 3.503 & 3.374 & 3.325 & \textbf{3.314} & 3.543 \\
AttnQ weight RMS & 1.01e-2 & 5.74e-3 & 8.44e-3 & 1.57e-2 & 2.95e-2 & 7.23e-2 \\
Mlp weight RMS & 1.25e-2 & 8.01e-3 & 1.27e-2 & 2.35e-2 & 4.51e-2 & 8.73e-2 \\
\bottomrule
\end{tabular}
\\ \footnotesize{\small *Except the first column, all other candidates are using Muon with controlled RMS.} 
\end{table}

\section{AdamW Baseline Scaling Law}
\label{sec:appendix:scaling}

To ensure the fairness and accuracy of our experiments, we conducted a series of experiments on our proprietary dataset to derive scaling law parameters that are optimal for AdamW. This includes determining the optimal model size($N$), number of training tokens($D$), learning rate($\eta$), batch size($B$) under a constrained computational budget (FLOPs, $C$).~\citep{kaplan2020scalinglawsneurallanguage,hoffmann2022trainingcomputeoptimallargelanguage,bi2024deepseek} Table \ref{tab:dense_scaling_param} presents the results of our systematic parameter search process.

\begin{table}[ht]
\small
\centering
\caption{Empirical Relationships Between Scaling Law Parameters and Computational Budget (FLOPs)}
\label{tab:dense_scaling_param}
\begin{tabular}{l c|c|c|c}
\toprule
 & $N(C)$ & $D(C)$ & $\eta(C)$ & $B(C)$ \\
\midrule
 & $0.0483359 \cdot C^{0.5112684}$ & $3.4480927 \cdot C^{0.4887316}$ & $0.0127339 \cdot C^{-0.0574752} $ & $0.0065202 \cdot C^{0.4137915}$ \\
\bottomrule
\end{tabular}
\end{table}

\paragraph{Hyper-Parameters Search} To systematically identify optimal scaling law hyper-parameters in the AdamW baseline, we adopted a multistage search protocol. First, we selected multiple computational budgets (FLOPs levels) and initialized model sizes, learning rates, and batch sizes based on empirical guidelines from prior studies. For each fixed FLOPs constraint, we varied the model size $N$ while adjusting the training token count $D$ inversely to maintain 
$C=6ND$, thereby exploring the trade-off between model capacity and data efficiency. Each configuration was trained to convergence, and the validation loss was recorded to determine the Pareto-optimal combinations of $N$ and $D$. Subsequently, with the optimal $N-D$ pairs fixed, we refined the learning rate and batch size through grid searches, ensuring stability and convergence across configurations. To mitigate local minima and enhance robustness, this iterative procedure was repeated 2–3 times, progressively narrowing the hyper-parameter space. 

\begin{figure}[t]
    \centering
    \includegraphics[width=0.95\textwidth]{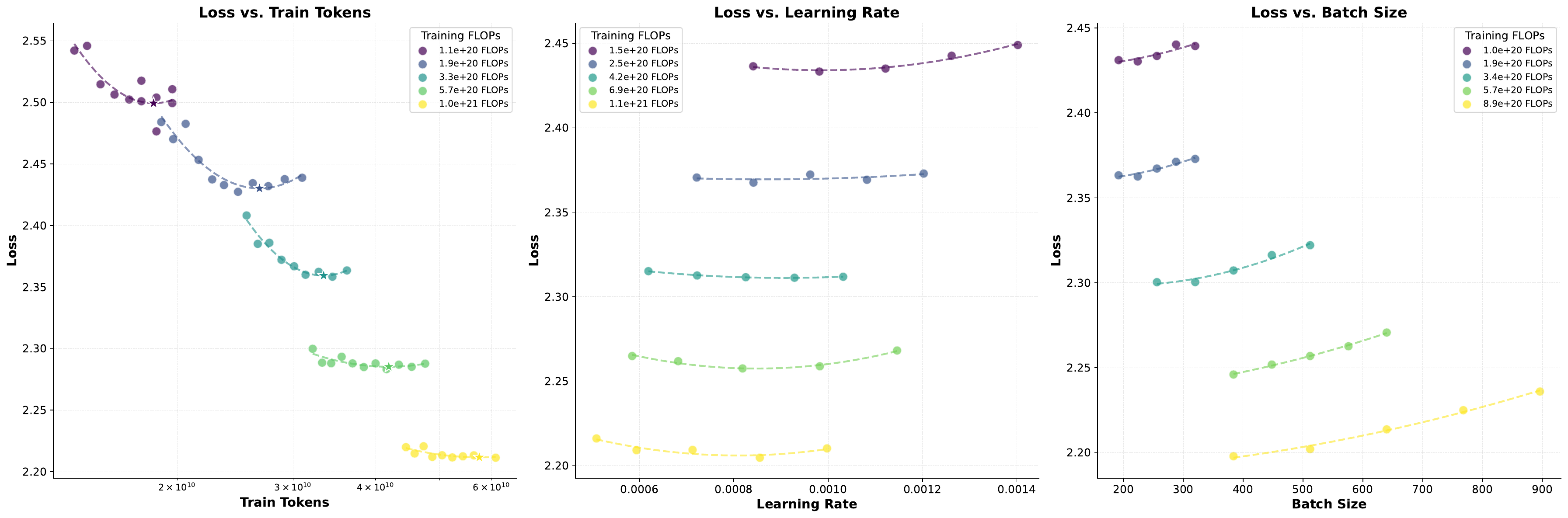}
    \caption{Optimization Landscapes for Scaling Law Hyper-parameters Across FLOPs Budgets} 
    \label{fig:scaling_search} 
\end{figure}

The optimization process is further illustrated in Figure \ref{fig:scaling_search}, which depicts the loss landscapes as functions of training tokens, learning rate, and batch size across varying FLOPs budgets. Each bowl-shaped curve represents the loss surface for a specific FLOPs level, with a distinct global minimum corresponding to the optimal hyper-parameter configuration. 

\section{Model Architecture}
\label{sec:appendix:modelarch}

Muon is agnostic to model architectures, and we used a model similar to Deepseek-V3-Small as described in \cite{deepseekai2024deepseekv3technicalreport}, because it is a strong model with open weights as a baseline. We made several small modifications in the Moonlight model and listed them here:

\paragraph{Multi-token Prediction (MTP)} MTP has not shown significant benefits to pretraining in our experiments. For simplicity, we do not introduce MTP layers into the Moonlight model.

\paragraph{Auxfree Bias Update} In \cite{deepseekai2024deepseekv3technicalreport}, auxfree bias is updated by: $b_i = b_i + u \times  \text{sign}(e_i)$, where $u$ is the update ratio, $b_i$ is the bias for the ith expert, and $e_i$ is the expert's violating ratio. We slightly modified the update rule as: $b_i = b_i + u \times (\text{sign}(e_i) - \text{sign}(e).\text{mean}())$, where $\text{sign}(e).\text{mean}()$ is the average of the signs of all expert's violating ratio, in order to control the magnitude of the bias, while does not change the topk selection logic.

\paragraph{Gate Scaling Factor} Deepseek-V2-Lite did not use the gate scaling factor, and Deepseek-V3 used a scaling factor of 2.5. We used a scaling factor of 2.446 to control a similar output rms like dense models. The code for calculating our gate scaling factor can be found in Figure \ref{fig:gate_scaling_code}.

\lstset{
  language=Python,                       
  backgroundcolor=\color{blue!5},        
  basicstyle=\footnotesize\ttfamily\color{black}, 
  keywordstyle=\color{blue!70}\bfseries,  
  commentstyle=\color{gray!70},          
  stringstyle=\color{red!80},             
  numberstyle=\tiny\color{blue!60},       
  stepnumber=1,                           
  numbersep=10pt,                         
  breaklines=true,                        
  showstringspaces=false,                 
  escapeinside={\%*}{*)},                 
  morekeywords={*,...}                    
}

\begin{figure}[h]
\begin{lstlisting}[frame=single,breaklines=false]
import numpy as np

def sigmoid(x):
    return 1 / (1 + np.exp(-x))

def calc_gate_scaling_factor(num_experts: int, topk: int, iter_times: int):
    """Calculate the gate scaling factor for MoE.

    Args:
        num_experts (int): The number of experts.
        topk (int): The number of experts to select.
        iter_timers (int): The number of iterations.

    Returns:
        float: The gate scaling factor.
    """
    factors = []
    for _ in range(iter_times):

        # mock gaussian logits
        logits = np.random.randn(num_experts)
        # select topk logits
        p = np.sort(sigmoid(logits))[::-1]
        p = p[:topk]
        # renormalize
        p = p / p.sum()
        # calculate the scaling factor
        factors.append( 1/ (p**2).sum()**0.5)
    return np.mean(factors)
\end{lstlisting}
\caption{Python implementation for calculating the gate scaling factor.}
\label{fig:gate_scaling_code}
\end{figure}

\section{Training Stability}
\label{sec:appendix:stability}

\begin{figure}[htbp]
    \centering
    \subfloat[Training Loss]{
        \includegraphics[width=0.48\textwidth]{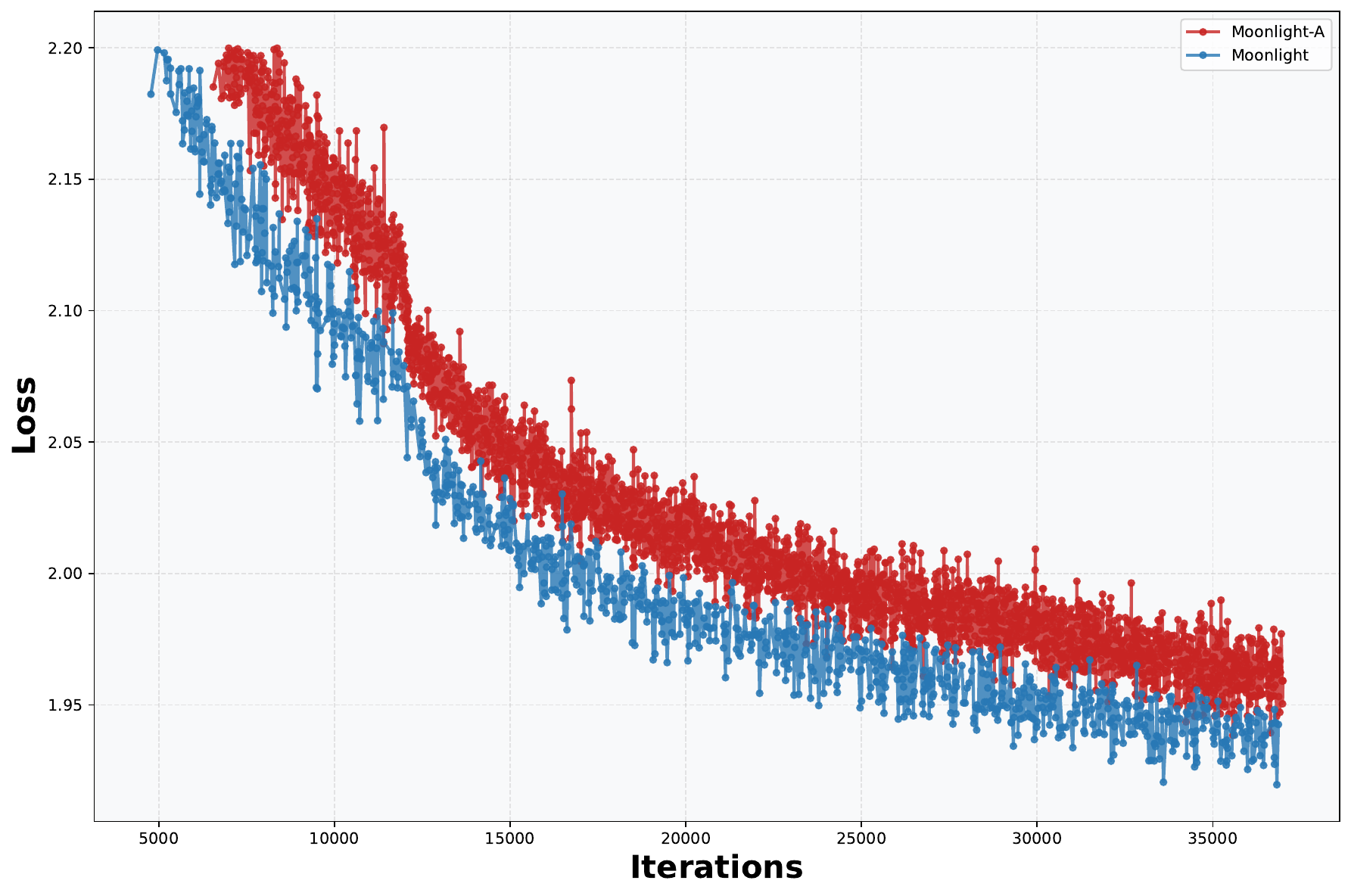}
    }
    \subfloat[Gradient Norm]{
        \includegraphics[width=0.48\textwidth]{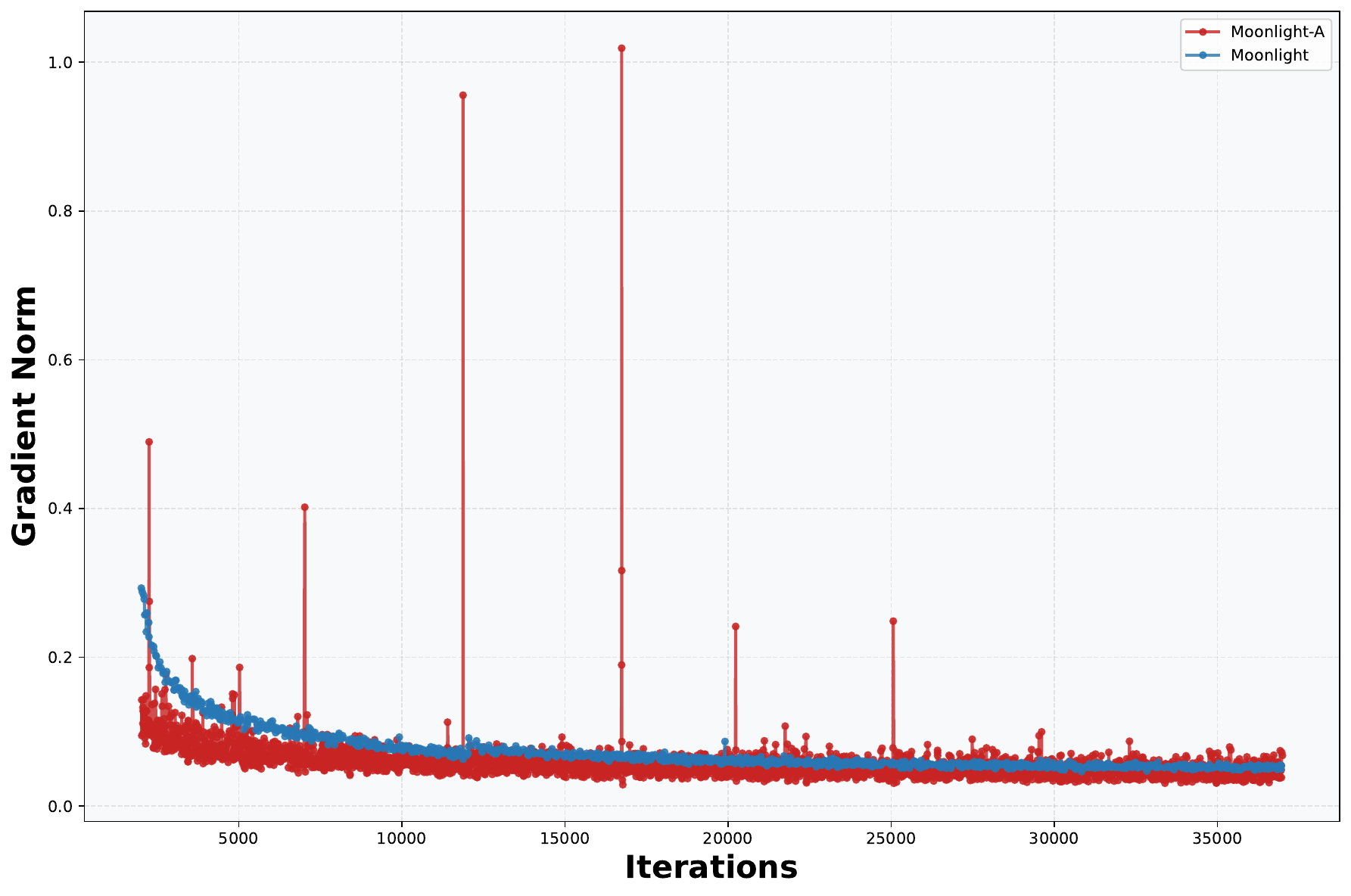}
    }
    
    \vspace{0.5em}
    
    \subfloat[Max Attention Logit (Layer 1)]{
        \includegraphics[width=0.48\textwidth]{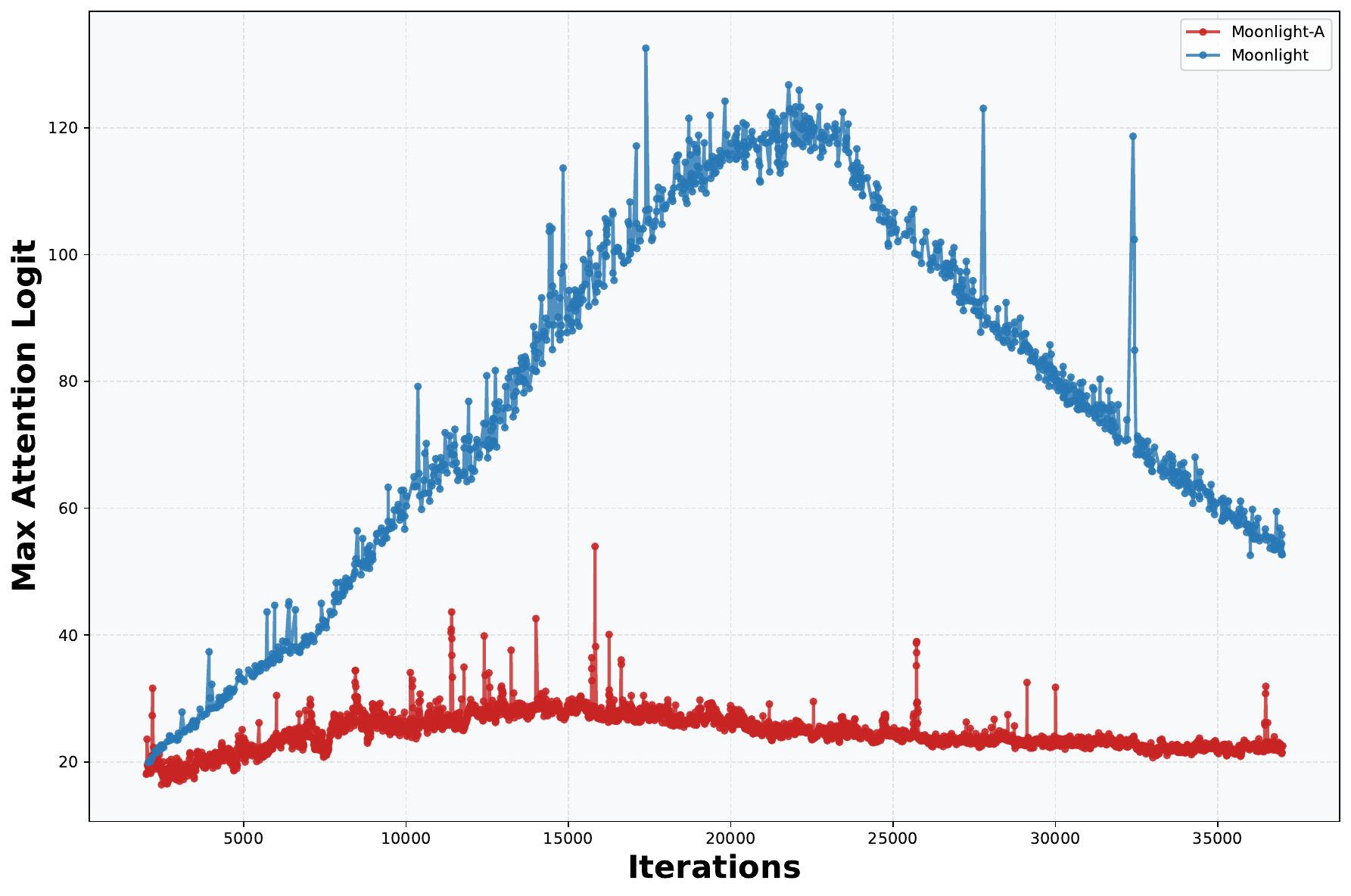}
    }
    \subfloat[Large Attention Logits Ratio (Layer 1)]{
        \includegraphics[width=0.48\textwidth]{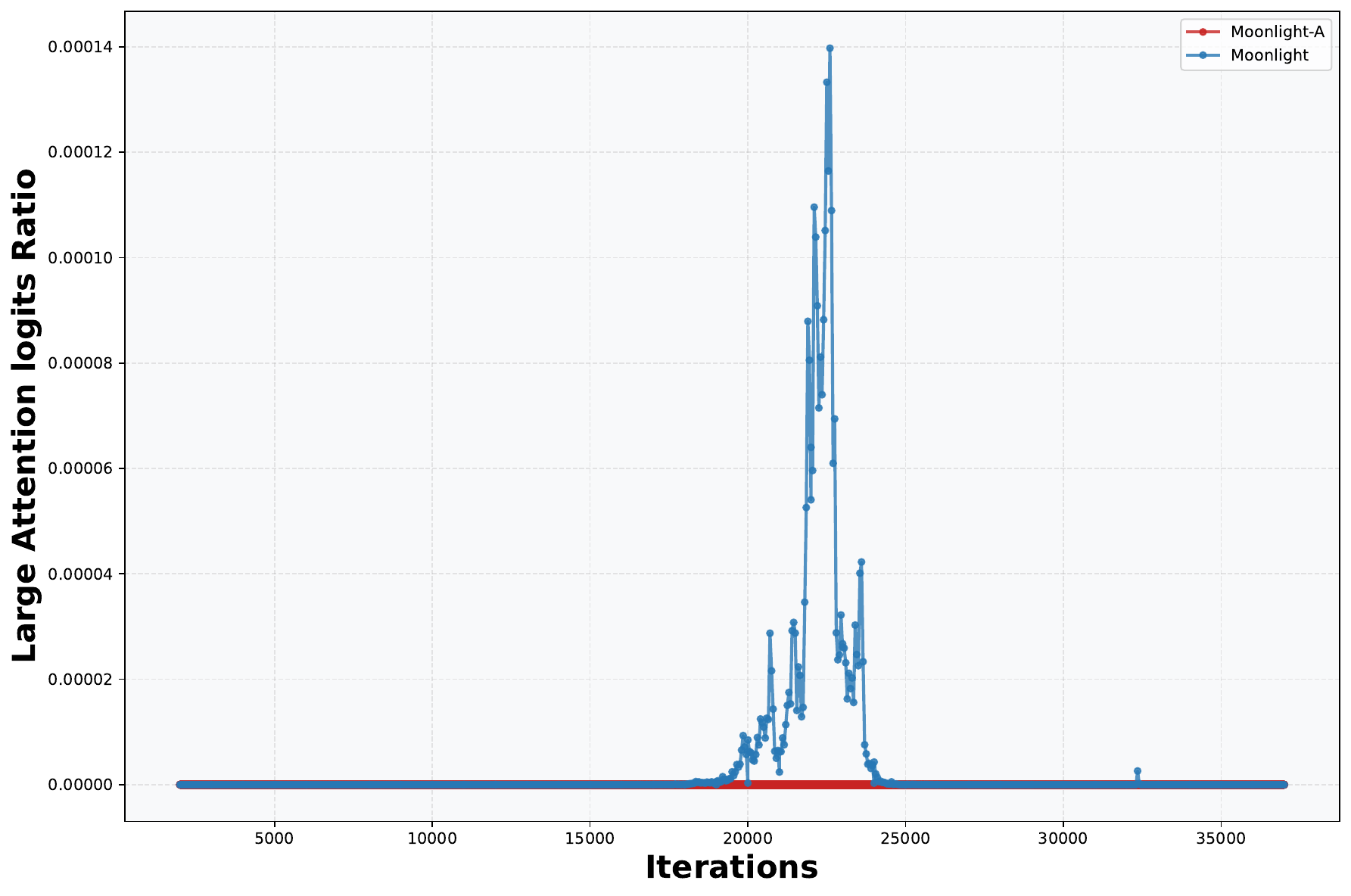}
    }
    
    \caption{Training dynamics comparison between Moonlight and Moonlight-A}
    \label{fig:training_dynamics}
\end{figure}

\paragraph{No Loss or Grad Norm Spike} The Moonlight training process was very smooth and we did not meet any loss spike or gradient norm spike. The loss and grad norm curve can be seen in Figure \ref{fig:training_dynamics} (Moonlight is colored in blue and Moonlight-A trained by AdamW is colored in red)

\paragraph{Max Attention Logit} During training, we observed that while both the training loss and gradient norm remained stable throughout the process, the maximum attention logit (computed as the single largest logit value across the global batch) exhibited a distinct upward trajectory in specific layers during the initial training phase, exceeding a threshold of 100. Notably, AdamW demonstrated healthier behavior in controlling this metric compared to alternative optimizers.

To further investigate the impacts of this phenomenon, we introduced the large attention logits ratio metric, defined as the proportion of attention logits exceeding 100 within a batch. As shown in Fig.\ref{fig:training_dynamics}, this ratio remained consistently low (about $10^{-4}$), indicating that extreme large logit values were sparse. Furthermore, the maximum logit values gradually decrease as training progressed, suggesting that the optimization dynamics become healthier.

\paragraph{RMSNorm Gamma Weight Decay} It is noteworthy that applying weight decay to the RMSNorm gamma parameter is crucial for ensuring training stability, as it effectively prevents excessively high output RMS values in each layer.

\section{Comparison with More Expensive Models}
\label{sec:appendix:comparisons}

\begin{figure}[t]
    \centering
    \includegraphics[width=0.8\textwidth]{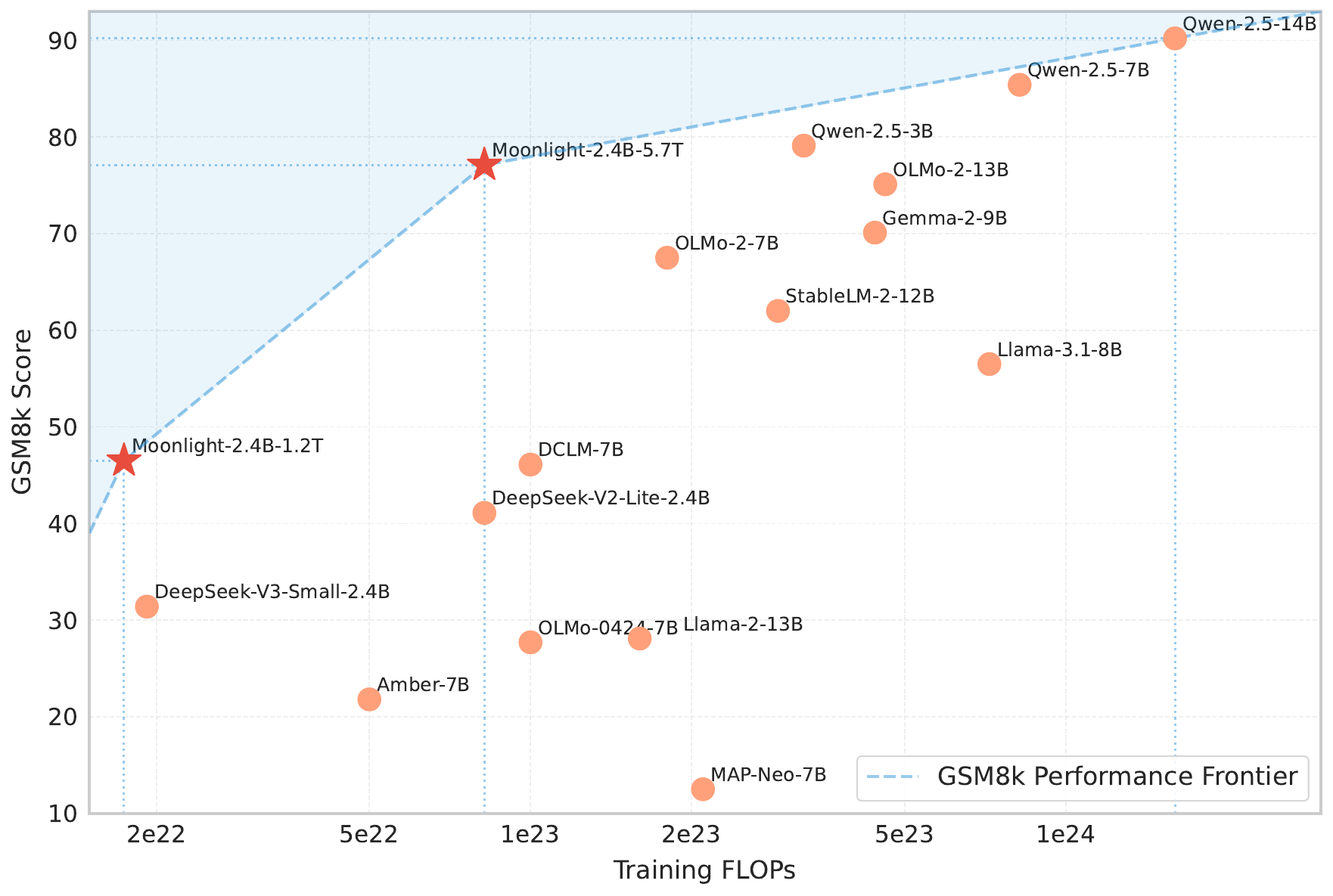}
    \label{fig:gsm8k}
    \caption{The GSM8k performance of our Moonlight model optimized with Muon and other comparable models.} 
    \label{Fig:model_perf} 
\end{figure}

Table \ref{tab:larger_compute} presents a comparative analysis between our Moonlight model (optimized with Muon) and publicly available models trained with greater computational resources, including \texttt{LLama3.1-8B}~\citep{grattafiori2024llama3herdmodels}, \texttt{Gemma-9B}~\citep{team2024gemma} and \texttt{Qwen2.5-7B}~\citep{qwen2.5}. Figure \ref{Fig:model_perf} illustrates the GSM8k performance benchmarks of Moonlight against comparable models in the field.

\begin{table}[!ht]
    \small
    \centering
    \caption{Comparison of different models on various benchmarks.}
    \setlength{\tabcolsep}{4pt}
    \begin{tabular}{@{}c l c || c c c@{}}
    \toprule
& \textbf{Benchmark (Metric)} & \textbf{Moonlight} & \textbf{LLAMA3.1-8B}  & \textbf{Gemma2-9B} & \textbf{Qwen2.5-7B}  \\
&  &   & \multicolumn{3}{c}{Larger Training Compute Model} \\
    \midrule
    & Activated Param$^{\dagger}$ & 2.24B & 7.38B & 8.32B & 6.83B \\
    & Total Params$^{\dagger}$   & 15.29B & 7.38B & 8.32B & 6.83B \\
    & Training Tokens & 5.7T & 15T & 8T & 18T \\
    & Optimizer  & Muon & AdamW & Unknown & Unknown \\
    \midrule
    \multirow{2}{*}{English} 
    & MMLU & 70.0 & 66.7 & 71.3 & 74.2  \\
    & MMLU-pro & 42.4 & 37.1 & 44.7 & 45.0 \\
    & BBH & 65.2 & 57.7 & 68.2 & 70.4 \\
    & TriviaQA$^{\ddagger}$ & 66.3 & 70.3 & - & 60.0 \\
    \midrule
    \multirow{2}{*}{Code} & HumanEval & 48.1 & 37.2 & 37.8 & 57.9 \\
    & MBPP & 63.8 & 47.6 & 62.2 & 74.9 \\
    \midrule
    \multirow{2}{*}{Math} & GSM8K & 77.4 & 57.2 & 70.7 & 85.4 \\
    & MATH & 45.3 & 20.3 & 37.7 & 49.8 \\
    \bottomrule
    \end{tabular}

    \footnotesize{\small$^{\dagger}$ The reported parameter counts exclude the embedding parameters.$^{\ddagger}$ We test all listed models with the full set of TriviaQA.} 
    \label{tab:larger_compute}
\end{table}

\section{Singular Value Distributions of Weight Matrices}
\label{sec:appendix:svd}
\begin{figure}[t]
    \centering
    \includegraphics[width=\textwidth]{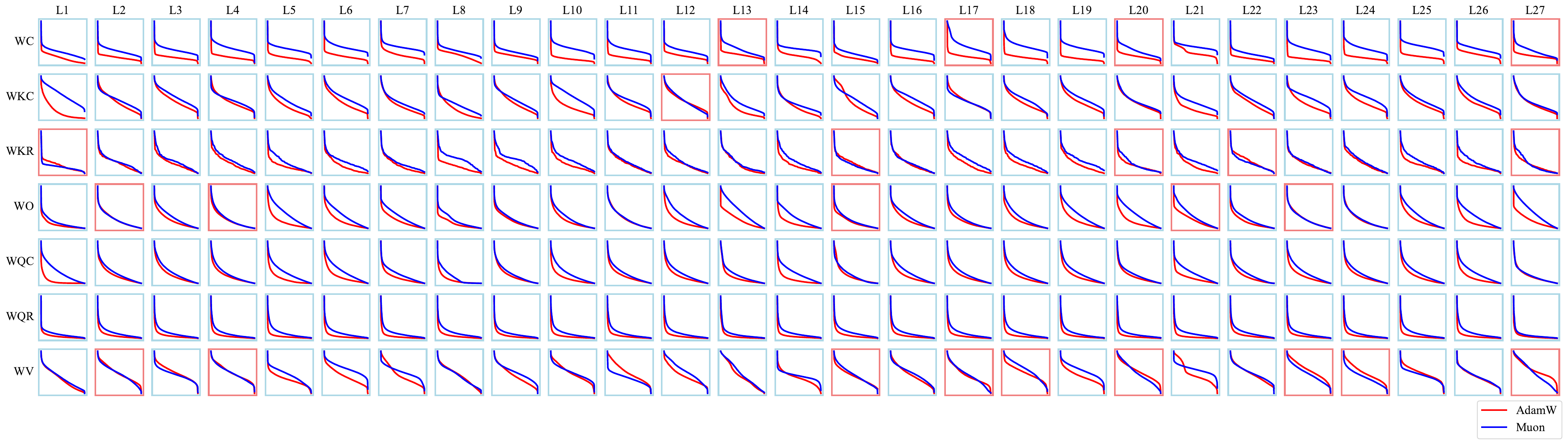}
    \caption{Distribution of singular values for each weight matrix in the attention layers. We use WC to denote the weight matrices at each layer that compress the hidden states to the shared latent spaces for keys and values, WV to denote the weight matrices up-projecting the values from the latent space, WO to denote the output projection matrices, and WKR, WKC, WQR and WQC to denote the projection matrices for the part of keys and queries with and without RoPE respectively. We set the spines of each line graph red if the corresponding weight matrix optimized by Muon has a lower singular entropy than AdamW.} 
    \label{fig_svd_attn} 
\end{figure}

\begin{figure}[t]
    \centering
    \includegraphics[width=\textwidth]{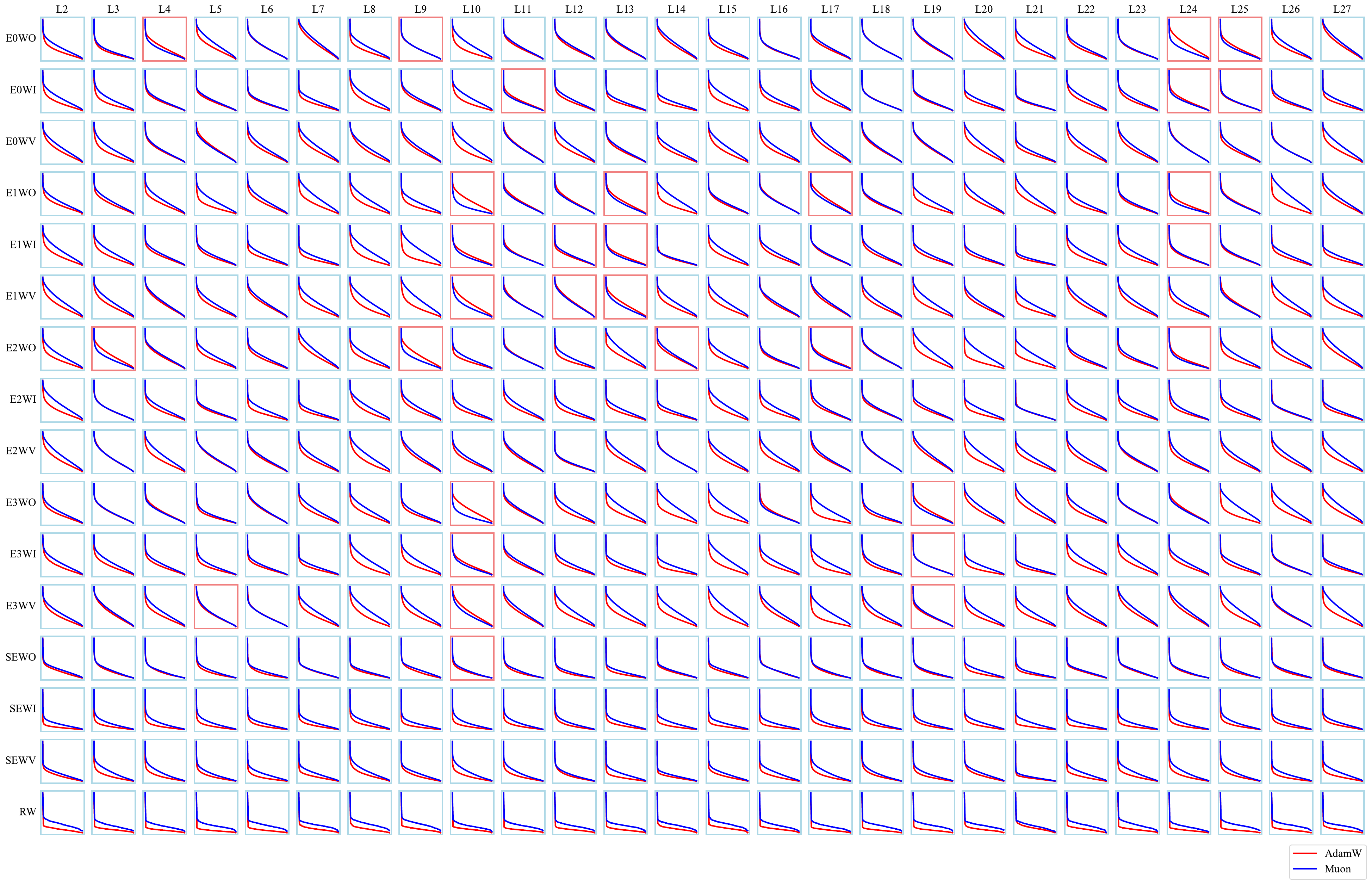}
    \caption{Distribution of singular values for each weight matrix in the feed-forward network (FFN) layers. We use WI, WV and WO to denote the weight matrices involved in the FFN layer with SwiGLU activation function, where WI represents the input projection to the Swish$_1$ function, WV represents the extra input projection interacting with Swish$_1$ activations, and WO represents the output projection. We use E0, E2, E3 to denote three arbitrarily selected expert models and SE to denote the weights in the shared expert model. We use RW to denote the weights in the router. We set the spines of each line graph red if the corresponding weight matrix optimized by Muon has a lower singular entropy than AdamW.} 
    \label{fig_svd_ffn} 
\end{figure}

We visualize the singular value distributions of weight matrices by plotting a line graph of its singular values in descending order for each matrix, normalized by the largest one. As shown in Figures \ref{fig_svd_attn} and \ref{fig_svd_ffn}, we find that, for most of the weight matrices, the singular value distributions of them optimized by Muon are more flattened than that of AdamW, which further confirms the hypothesis that Muon can provide a more diverse spectrum of updates.

\end{document}